# AgentCoMa: A Compositional Benchmark Mixing Commonsense and Mathematical Reasoning in Real-World Scenarios

**Lisa Alazraki**[†], **Lihu Chen**[†], **Ana Brassard**[♭], **Joe Stacey**[§], **Hossein A. Rahmani**[‡], **Marek Rei**[†]
[†]Imperial College London, [♭]RIKEN, [§]University of Sheffield, [‡]University College London
{lisa.alazraki20, lihu.chen, marek.rei}@imperial.ac.uk
ana.brassard@riken.jp, j.stacey@sheffield.ac.uk, hossein.rahmani.22@ucl.ac.uk

## Abstract

Large Language Models (LLMs) have achieved high accuracy on complex commonsense and mathematical problems that involve the composition of multiple reasoning steps. However, current compositional benchmarks testing these skills tend to focus on *either* commonsense or math reasoning, whereas LLM agents solving real-world tasks would require a combination of *both*. In this work, we introduce an **Agent**ic **Co**mmonsense and **Ma**th benchmark (AgentCoMa), where each compositional task requires a commonsense reasoning step *and* a math reasoning step. We test it on 61 LLMs of different sizes, model families, and training strategies. We find that LLMs can usually solve both steps in isolation, yet their accuracy drops by nearly 30% on average when the two are combined. This is a substantially greater performance gap than the one we observe in prior compositional benchmarks that combine multiple steps of the same reasoning type. In contrast, non-expert human annotators can solve the compositional questions and the individual steps in AgentCoMa with similarly high accuracy. Furthermore, we conduct a series of interpretability studies to better understand the performance gap, examining neuron patterns, attention maps and membership inference. Our work underscores a substantial degree of model brittleness in the context of mixed-type compositional reasoning and offers a test bed for future improvement.

## 1 Introduction

LLM agents performing real-world tasks must be able to combine different types of reasoning (Furuta et al., 2024; Ma et al., 2025). For example, an agent planning the weekly groceries for a vegetarian user may use commonsense to identify suitable items, and mathematical reasoning to continuously track costs and ensure the total stays within budget. Indeed, the performance of LLMs in reasoning-intensive tasks has been rapidly improving, reaching high accuracy on commonsense benchmarks that test the ability to reason over everyday situations (Sap et al., 2020; Bhargava and Ng, 2022) as well as mathematical ones testing logical and systematic thinking (Wang et al., 2026; Yan et al., 2025). Although these results are impressive and potentially useful for specific applications (Yen and Hsu, 2023; Zhao et al., 2023), they are obtained on self-contained questions that do not deliberately combine multiple types of reasoning. Hence, they do not fully reflect the difficulty of real-world tasks.

At the same time, existing agentic benchmarks do not exclusively focus on combining different reasoning types. Instead, they introduce further elements of difficulty such as tool calling, long task horizons, or the necessity to adapt to a constantly changing environment (Yehudai et al., 2025a). This makes them unsuitable for evaluating the mixed-type compositional reasoning skills of LLMs in a controlled manner.

Motivated by the lack of datasets for systematically evaluating the ability of LLMs to combine different types of reasoning in agentic settings, we introduce AgentCoMa, a compositional benchmark where each question requires both a commonsense reasoning step and a mathematical reasoning step. We choose commonsense and math as prior literature presents them as dissimilar yet complementary types of reasoning (Davis, 2023; Ziabari et al., 2025): commonsense is fast and intuitive, akin to Kahneman (2011)'s System 1 reasoning, while math is slow and deliberative and corresponds to System 2 reasoning. We ground the questions in AgentCoMa in five real-world agentic scenarios: *house working*, *web shopping*, *science experiments*, *smart assistant* and *travel agent*.

We test and analyse the performance of 61 contemporary LLMs on AgentCoMa. These belong to

Our data, code and leaderboard submission form can be accessed from `https://agentcoma.github.io`.

different model families, with sizes ranging from 1.5B to 141B, and include instruction-tuned LLMs, mixture-of-experts (MoE) models, as well as LLMs optimised for reasoning via both supervised fine-tuning and reinforcement learning. We observe that the compositionality gap (Press et al., 2023)—i.e., the difference between the compositional accuracy and the proportion of samples where *both* steps are answered correctly when attempted in isolation—is high: while the 61 LLMs achieve a median accuracy above 85% on each step and can independently solve both in nearly 75% of cases, their median accuracy on the compositional task is only 42%. Note that non-expert human annotators are able to solve both reasoning steps in isolation approximately as often as the compositional questions.

In contrast, we find that for established compositional benchmarks where both reasoning steps are of the same type—i.e., knowledge-based Bamboogle (Press et al., 2023) and math-based MultiArith (Roy and Roth, 2015)—the compositionality gap is low or non-existent, and failure to reach the correct solution for a compositional question is mainly due to the model's inability to answer correctly one or more of its underlying sub-questions.

To understand *why* the questions in AgentCoMa are challenging for LLMs despite containing 'easy' reasoning steps, we analyse context length, internal knowledge mechanisms and attention patterns under different types of reasoning, as well as the presence of similar tasks in the training data. We find that mixed-type reasoning questions are a relatively unobserved pattern for LLMs, which may explain why the models tend to activate neurons relevant to only one reasoning type (math) when solving the compositional task, leading to a performance collapse.

In summary, our contributions are:

1. We introduce AgentCoMa, a benchmark of high-quality, human-written questions set in agentic scenarios, which require both commonsense and mathematical reasoning to be solved. To the best of our knowledge, ours is the first dataset for the systematic evaluation of mixed-type compositional reasoning in LLMs.

2. We benchmark 61 contemporary LLMs on AgentCoMa, and observe large compositionality gaps for all of them. In contrast, no significant gap is observed when the same questions are answered by human annotators, and LLMs solving existing compositional benchmarks achieve negligible compositionality gaps.

3. Our analysis shows that mixed-type compositional reasoning questions are relatively rare in LLM training data, hence the models fail to leverage the correct neurons when answering them. Instead, they revert to learned neural circuits associated with a single reasoning type. AgentCoMa thus uncovers a fundamental brittleness in LLMs, and provides a platform for evaluating future advancements.

## 2 Related Work

**Commonsense reasoning** in language models has been thoroughly investigated, with over a hundred benchmarks developed in recent years for this purpose (Davis, 2023). Current decoder-based LLMs have been shown to achieve high accuracy (Brown et al., 2020; Chowdhery et al., 2023; Anil et al., 2023; Wu et al., 2023; Achiam et al., 2024) on several existing commonsense tasks (Kavumba et al., 2019; Talmor et al., 2021; Gupta et al., 2023; Li et al., 2025). As discussed by Davis (2023), commonsense comprises several domains, including temporal and spatial reasoning (Zhou et al., 2019; Aroca-Ouellette et al., 2021; Qin et al., 2021; Liu et al., 2022), reasoning over social relations (Sap et al., 2019), and causal reasoning in everyday situations (Gordon et al., 2012; Ghosal et al., 2021).

**Mathematical reasoning** has seen a surge of interest in recent years (Lu et al., 2023), with the creation of several new benchmarks. These range from math word problems involving elementary calculations (Roy and Roth, 2015; Koncel-Kedziorski et al., 2016; Miao et al., 2020; Cobbe et al., 2021; Patel et al., 2021), on which contemporary LLMs achieve high accuracy (Forootani, 2025), to more challenging big-number arithmetic questions which require tool-augmentation to be solved (Gao et al., 2023; Hao et al., 2023; Alazraki and Rei, 2025), to competition-level problems on advanced topics such as number theory, combinatorics and calculus, on which LLMs score consistently below 50% (Glazer et al., 2024; He et al., 2024; Zhong et al., 2024; Gao et al., 2025). It is worth noting that the math reasoning steps in AgentCoMa are elementary, i.e., similar in difficulty to those in Roy and Roth (2015) and Cobbe et al. (2021).

**Compositional reasoning** involves analysing com-plex problems by isolating their basic elements and then combining those elements to generate new inferences or solutions (Shi et al., 2025); a skill that is innately human (Lake et al., 2019) and fundamental for generalisable learning (Hupkes et al., 2023). Compositional tasks contain multiple reasoning steps, each relying on the result of the previous one. Individual steps may involve any reasoning type (e.g. commonsense, math), and they should be easy when solved in isolation, so that compositional ability is the sole focus of the evaluation (Xu et al., 2024; Zhao et al., 2025). While AgentCoMa deliberately combines different types of reasoning, in existing compositional benchmarks all steps are usually of the same type, e.g. commonsense (Geva et al., 2021; Zhan et al., 2026) or math (Cobbe et al., 2021; Patel et al., 2021).

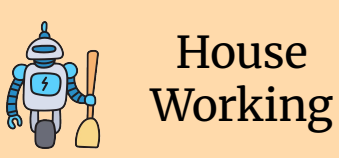

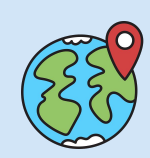

Figure 1: Question domains in AgentCoMa with example questions from the development set.

**Agentic reasoning** in LLMs is defined as the ability to simultaneously reason and act in the world (Plaat et al., 2026). LLM agents are often equipped with tools—i.e., executable programs or APIs—that they can call on to interact with their environment (Parisi et al., 2022). Tool use is however only one aspect of agentic reasoning: in order to use tools effectively, agents need to be able to plan through multiple time steps (Gui et al., 2025) and reason over complex constraints and aspects of the world (Christakopoulou et al., 2024; Kim et al., 2025). Hence, agentic behaviour is intrinsically compositional and requires a diverse range of reasoning skills, as already noted by Furuta et al. (2024); Rasal (2024). Existing agentic benchmarks are primarily focused on tool use (Deng et al., 2024; Guo et al., 2024; Liu et al., 2024), multi-agent cooperation (Smith et al., 2024; Chang et al., 2025), growing task horizons (Kwa et al., 2025; Wijk et al., 2025; Alazraki et al., 2026) and dynamic environments (Paglieri et al., 2025; Park et al., 2026). Our aim is instead to test the compositional abilities of LLMs in real-world tasks that require diverse types of reasoning, in the absence of other confounding factors. Identifying the failure modes of LLMs in such a controlled setting can inform future strategies for enhancing their reasoning.

## 3 The AgentCoMa Benchmark

### 3.1 Task Desiderata

We aim to create agentic questions grounded in the domains shown in Figure 1. We choose these domains as they represent useful real-world applications of LLM agents (Chen et al., 2024, 2025b; Mok et al., 2025). Each question combines a commonsense reasoning step and a mathematical reasoning step. Following Davis (2023), the commonsense steps are designed to be easy for humans, require rich knowledge and complex reasoning, be relevant to end-user AI tasks, and have clear-cut criteria of correctness. Similarly to prior benchmarks (Luo et al., 2025), we accept commonsense steps that are based on common knowledge (e.g., *France is in Europe*). Each mathematical reasoning step must involve one single arithmetic operation. In Section 3.3, we elaborate on the validation process for verifying the above criteria.

Each question in AgentCoMa is manually created by expert annotators, given instructions shown in Appendix H, and without the aid of LLMs or other automated processes. Preliminary experiments with GPT-4o[1] showed that even the most capable LLMs are unable to write questions similar to those in AgentCoMa, including when prompted with similar ones in-context (details of this preliminary experiment can be found in Appendix B).

| | Question | Answer |
|---|---|---|
| Composition | You need to organize the tools in the garage. You have 1 power drill, 6 hammers, 3 extension cords, 1 leaf blower, and 7 screwdrivers. You need to store all electrical items in a weatherproof cabinet. How many items go into the weatherproof cabinet? | 5 |
| Commonsense | You need to organize the tools in the garage. You have a power drill, hammers, extension cords, a leaf blower, and screwdrivers. You need to store all electrical items in a weatherproof cabinet. Which items go into the weatherproof cabinet? | power drill, extension cords and leaf blower |
| Math | You need to organize the tools in the garage. You have 1 power drill, 6 hammers, 3 extension cords, 1 leaf blower, and 7 screwdrivers. You need to store the power drill, the extension cords and the leaf blower in a weatherproof cabinet. How many items go into the weatherproof cabinet? | 5 |

Figure 2: A data sample in AgentCoMa. The compositional question requires a choice among multiple options based on commonsense, followed by an arithmetic step. The commonsense sub-question is a *which* question only requiring the commonsense-based choice. In the math sub-question, the solution to the commonsense step is given in the question, and only arithmetic is required.

| **Validation step** | **Completed by** | |
|---|---|---|
| | **Sample author** | **Other expert** |
| Is the sample directed to an AI agent, requiring to solve a task with real-world utility? [Yes / No] | ✓ | ✗ |
| Does the commonsense portion require everyday knowledge beyond what can be inferred from the question? [Yes / No] | ✗ | ✓ |
| Does the question require a choice among multiple items based on commonsense, followed by a single arithmetic operation? [Yes / No] | ✗ | ✓ |
| Solve the sample and provide the final answer. | ✗ | ✓ |
| Describe ambiguities or other issues with the sample (if any). | ✗ | ✓ |

Table 1: Sample validation steps for AgentCoMa. Only samples that have successfully passed all steps are included in the benchmark.

### 3.2 Task Structure

The questions in AgentCoMa comprise two steps. The first requires choosing between multiple options based on commonsense. For example, given several food items with different prices, an agent has to decide which to buy subject to the constraint that purchased items must be safe to store at room temperature. The second step requires performing arithmetic, e.g., computing the total price of the cart.

Note that for each compositional question we also provide its two reasoning steps as distinct sub-questions. Therefore, each data sample consists of a compositional question, the corresponding commonsense reasoning and math reasoning sub-questions, and their respective ground-truth answers. Figure 2 illustrates a sample from the development set.

### 3.3 Data Validation

Each data sample in AgentCoMa has undergone a multi-step validation process, shown in Table 1, before being included in the dataset. The samples are split into distinct portions, each assigned to one of five expert annotators. Firstly, a sample is assessed via a binary *yes/no* questionnaire. In case of a negative answer to any of the binary questions, the sample is rewritten and the validation repeated. Further, the experts are tasked with solving each sample, and the answers are compared with the ground-truth answers. If a mismatch between the answers is found, the sample is rectified or rewritten and must be validated again. At this stage, experts can also explicitly point out ambiguities or other issues within the sample. Note that, to minimise potential bias, all but one of the assessment steps must be completed by an expert *other* than the one who authored the sample.

### 3.4 Data Statistics

AgentCoMa comprises 260 total samples, split into an 80-sample development set and a 180-sample test set. As the benchmark is meant for the evaluation of pre-trained LLMs, we do not provide a training set. The number of testing samples is consistent with established compositional benchmarks, e.g. Bamboogle (Press et al., 2023) (125 samples) and MultiArith (Roy and Roth, 2015) (180 samples). In both data splits, the samples are evenly distributed among the five agentic domains shown in Figure 1, and within each domain the four arithmetic operations (addition, subtraction, multiplication and division) are equally represented. Note that all models in Section 5 are benchmarked on the test set. Dev set results are in Appendix A.

[1] https://platform.openai.com/docs/models/gpt-4o

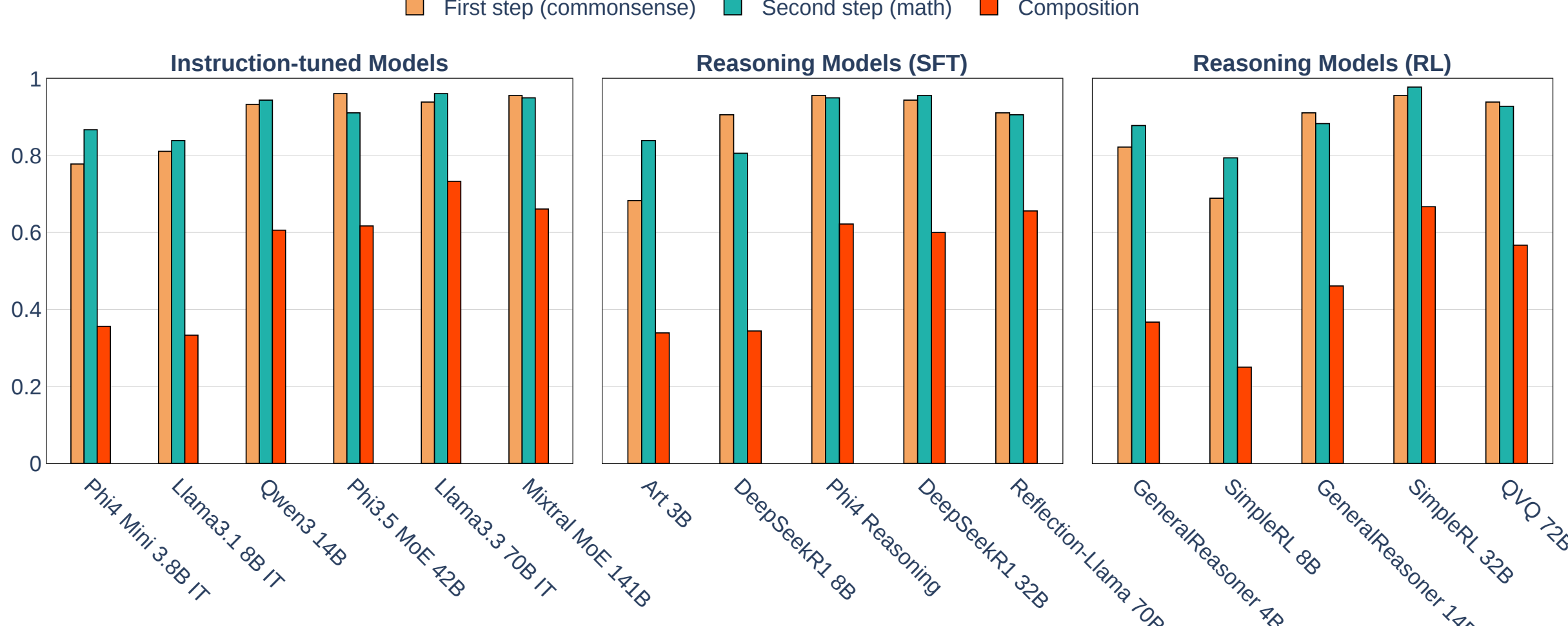

Figure 3: Accuracies obtained on AgentCoMa by 16 LLMs, grouped by tuning strategy and ordered by increasing model size. All models display a considerable performance gap between the individual reasoning steps (left-most and centre bars, in yellow and green) and their composition (right-most bars, in red). We run each LLM once on the entire test set using deterministic (greedy) sampling.

## 4 Experimental setup

### 4.1 Models

We benchmark 61 LLMs in total, and select from them 16 recently-released models to show representative results in the next sections. These have sizes ranging from 3B to 141B, and comprise six instruction-tuned LLMs (two of which rely on MoE architectures), five reasoning LLMs optimised via supervised fine-tuning (SFT), and a further five reasoning LLMs optimised via online reinforcement learning (RL). Model names and sizes are shown in Table 2. We include the results for all 61 LLMs in Appendix A, with additional results for API models in Appendix F. Model identifiers and inference hyperparameters are given in Appendix D.

### 4.2 Inference and Evaluation

LLM responses for all questions and sub-questions are elicited via few-shot chain-of-thought (CoT) prompting (Wei et al., 2022), with two in-context examples (shown in Appendix E.1), using greedy decoding in all cases. All responses are evaluated for accuracy. For questions requiring a numerical answer (i.e., compositional questions and math sub-questions), we extract the final number from the CoT response via regular expressions and check that it exactly matches the ground truth. Questions requiring non-numerical answers (i.e., commonsense sub-questions) are labelled as 'correct' or 'incorrect' by an LLM-as-a-judge (Zheng et al., 2023) given the ground-truth. This method has been shown to correlate highly with human assessment both in prior work (Huidrom and Belz, 2025; Zhou et al., 2025) and by our own experiments (see Appendix G). The LLM-as-a-judge prompt is shown in Appendix E.3.

### 4.3 Human Performance Study

We compare LLM results against human performance by carrying out a human study on the entire test set of AgentCoMa. We employ 45 crowdworkers and have them solve both the composi-tional questions and the sub-questions. Each crowd-worker evaluates a distinct set of 12 questions. To prevent answer contamination, sets are constructed so that the same crowd-worker will not be assigned questions and sub-questions from the same data sample. Crowd-workers are non-experts, with the only requirement being high-school-level education and fluency in English. They are required to solve each question by hand without using tools such as calculators or web search. Further details and annotator guidelines are given in Appendix I.

| Model Name | Model Type | First step | Second step | Both correct | Comp. |
|---|---|---|---|---|---|
| Phi4 Mini 3.8B IT | *Instruction-tuned* | 77.8 | 86.7 | 66.1 | 35.6 |
| Llama3.1 8B IT | | 81.1 | 83.9 | 68.3 | 33.3 |
| Qwen3 14B | | 93.3 | 94.4 | 88.9 | 60.6 |
| Phi3.5 MoE 42B IT | | 96.1 | 91.1 | 87.2 | 61.7 |
| Llama3.3 70B IT | | 93.9 | 96.1 | 90.0 | 73.3 |
| Mixtral MoE 141B | | 95.6 | 95.0 | 90.6 | 66.1 |
| Art 3B | *Reasoning (SFT)* | 68.3 | 83.9 | 57.8 | 33.9 |
| DeepSeekR1 8B | | 90.6 | 80.6 | 72.2 | 34.4 |
| Phi4 Reasoning 14.7B | | 95.6 | 95.0 | 91.7 | 62.2 |
| DeepSeekR1 32B | | 94.4 | 95.6 | 90.0 | 60.0 |
| Reflection-Llama 70B | | 91.1 | 90.6 | 82.2 | 65.6 |
| GeneralReasoner 4B | *Reasoning (RL)* | 82.2 | 87.8 | 73.9 | 36.7 |
| SimpleRL 8B | | 68.9 | 79.4 | 56.7 | 25.0 |
| GeneralReasoner 14B | | 91.1 | 88.3 | 80.0 | 46.1 |
| SimpleRL 32B | | 95.6 | 97.8 | 93.9 | 66.7 |
| QVQ 72B | | 93.9 | 92.8 | 87.8 | 56.7 |
| Non-expert human | – | 84.4 | 89.4 | 78.9 | 82.8 |

Table 2: Fine-grained AgentCoMa accuracies. We show the proportions of questions where *both* steps succeed individually ('both correct'). We contrast this with the observed compositional accuracy ('comp.').

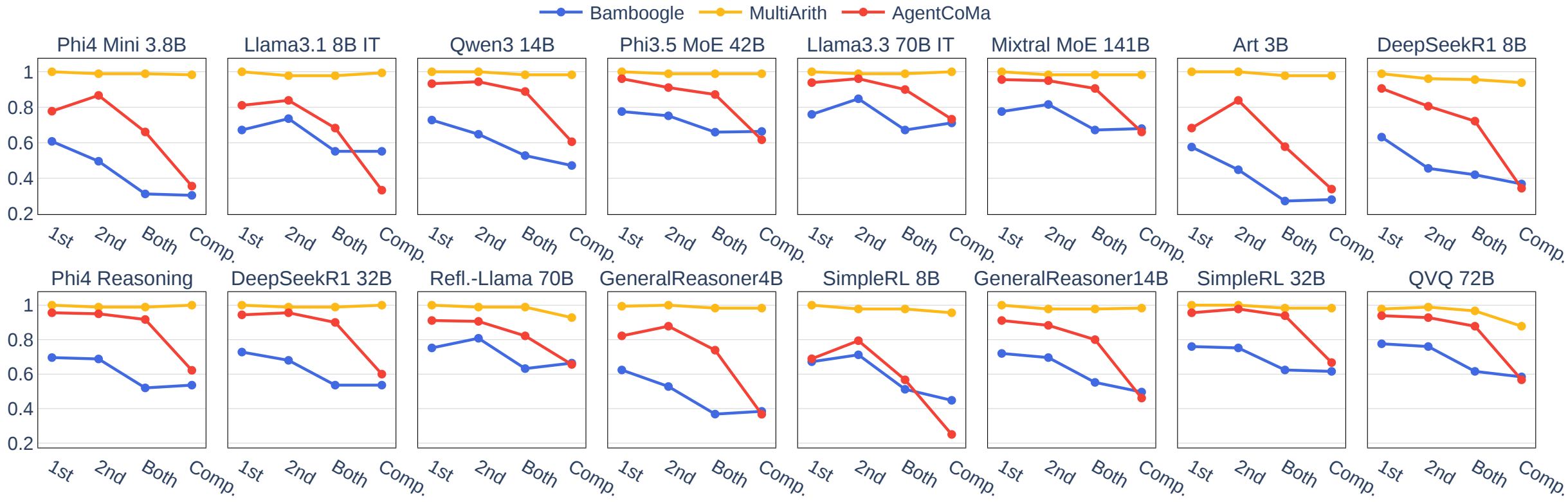

Figure 4: Accuracies on AgentCoMa, Bamboogle and MultiArith. We evaluate on either ('1st', '2nd') and both steps solved in isolation, and their composition ('comp.'). The compositionality gap, corresponding to the slope of the line segment between 'both' and 'comp.', is large for AgentCoMa but small for Bamboogle and MultiArith.

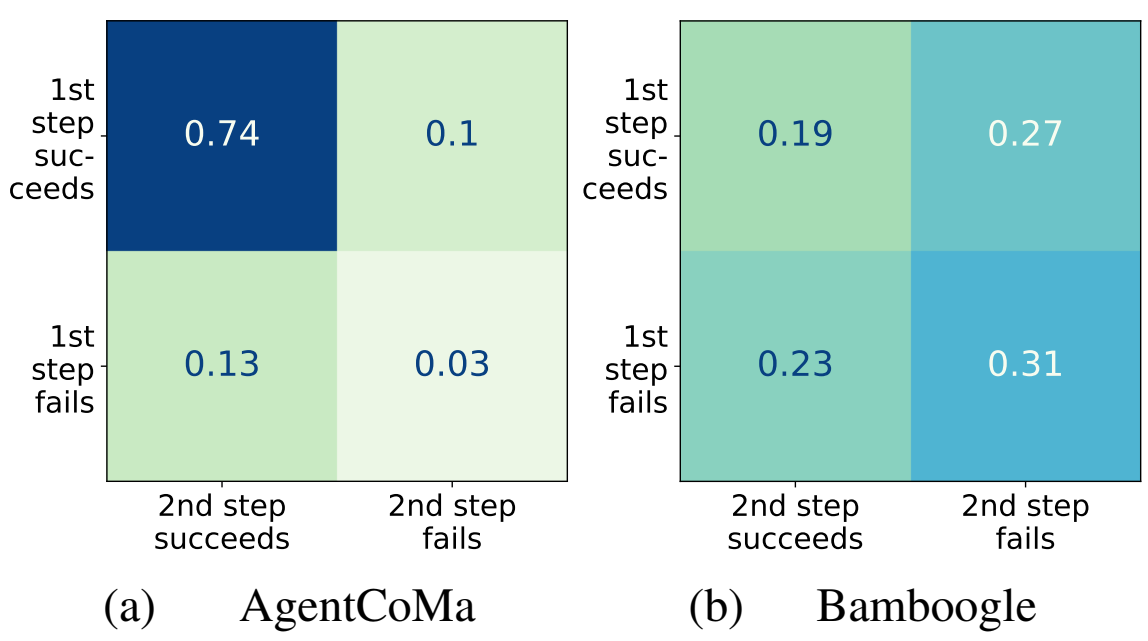

Figure 5: Proportions of failed compositional questions in AgentCoMa (a) and Bamboogle (b). We group failures according to whether the same LLM *succeeds* or *fails* on each of the underlying reasoning steps. Proportions are averaged across all 16 LLMs shown in Table 2.

## 5 Results

As shown in Figure 3, all LLMs achieve substantially higher accuracies on the individual sub-questions—i.e., first step (commonsense) and second step (math)—than on their composition. Table 2 shows that the models can usually solve correctly *both* steps in a sample when these are presented in isolation (avg. 80%), yet their mean compositional accuracy is only 51%, resulting in a 29% average compositionality gap. Note that we also experiment with question decomposition strategies other that CoT, and find that they lead to similarly low performance (see Appendix C for details).

Notably, the compositionality gap is as large for reasoning models, including RL-tuned models, as it is for instruction-tuned models. Hence, despite recent literature showing that RL reasoning optimisation allows LLMs to generalise well to a variety of out-of-distribution tasks (Huan et al., 2025), our experiments show that even these sophisticated models struggle with mixed-type compositional reasoning, performing as poorly as models tuned via imitation learning. In contrast, we observe that non-expert humans can solve both individual steps in a sample approximately as often as they can solve the compositional task[2], as shown in Table 2.

As a result of the compositionality gap, most LLM compositional failures on AgentCoMa happen when the same model can correctly solve both reasoning steps in isolation. We compute the proportion of compositional failures for each model according to whether they succeed or fail on each of the underlying steps. These proportions, averaged across all LLMs, are illustrated in Figure 5(a). We observe that in approximately three-quarters of failures, the model succeeds at both individual steps. This proportion is even higher when we only consider the larger, more capable LLMs (see Appendix L.1). We include example outputs where the individual steps succeed but their composition fails in Appendix L.2.

[2]Note that the compositional question and each sub-question within a sample are always attempted by *different* non-expert annotators. This may explain why the joint correctness of the sub-questions is slightly less frequent.

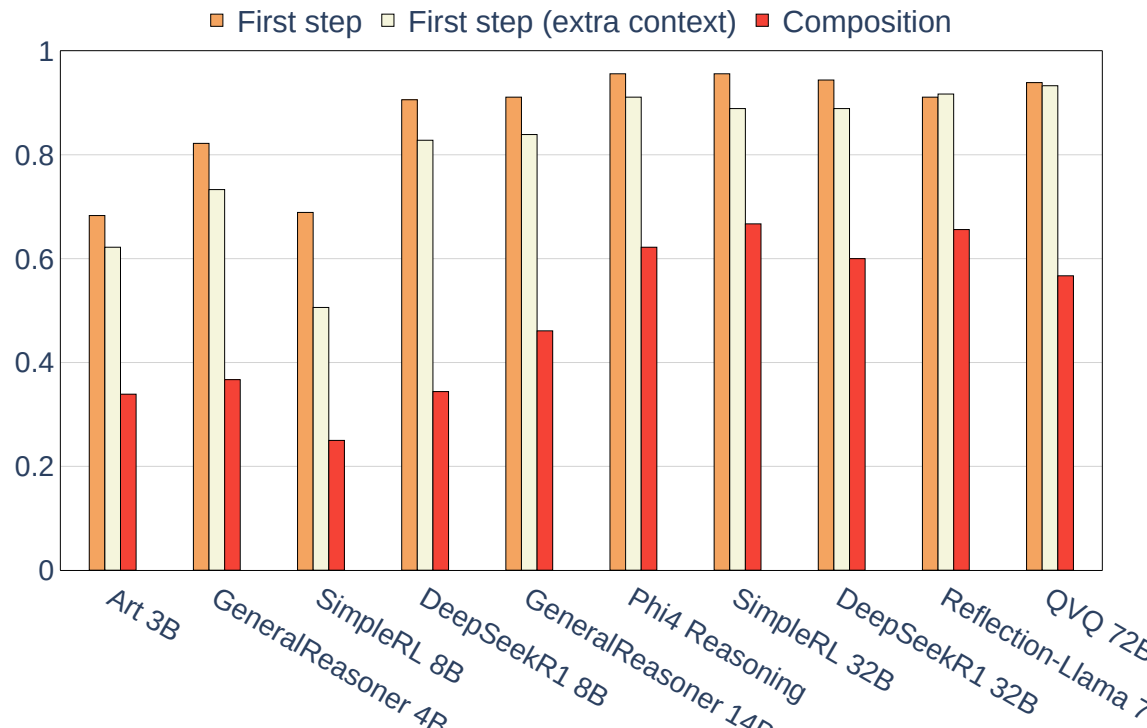

Figure 6: Accuracies of all reasoning models on first-step sub-questions with extra context, compared to the same sub-questions without extra context and the compositional questions. Note that instruction-tuned models follow a similar trend, as shown in Appendix K.

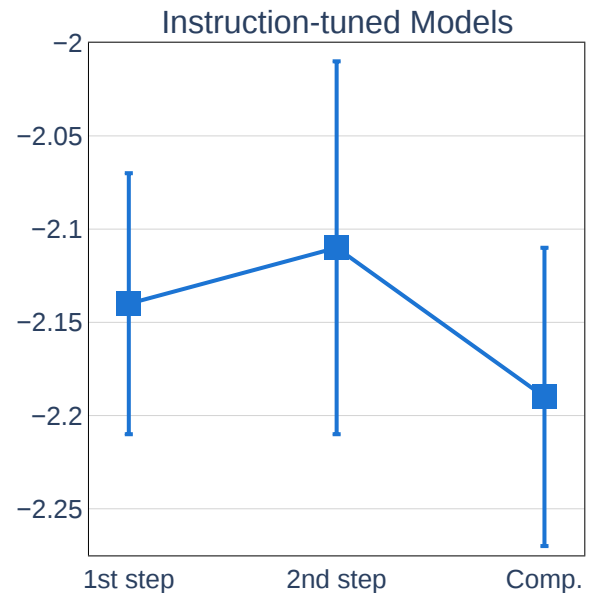

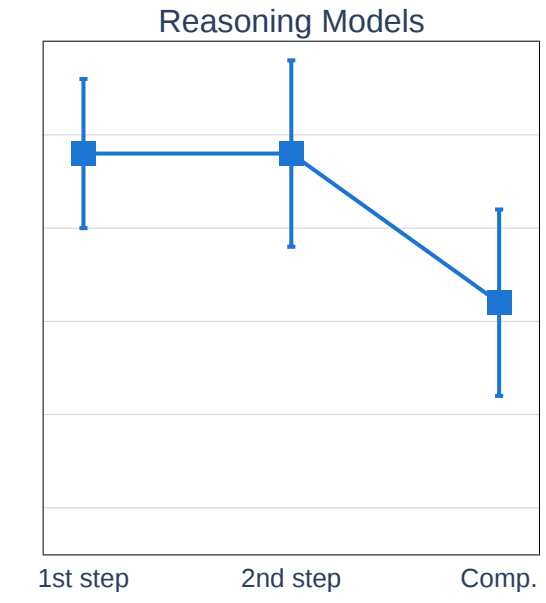

Figure 7: Average Min-K%++ scores for each individual reasoning step and the compositional questions ('comp.') in AgentCoMa, computed for all instruction-tuned models (left) and all reasoning-optimised models (right). In both cases, the lower Min-K%++ shows that the mixed-type reasoning task represents a relatively unseen pattern in the LLMs training data.

### 5.1 Comparison with Other Benchmarks

We compare LLM performance on AgentCoMa with two established compositional benchmarks where the reasoning steps are of the same type: Bamboogle (Press et al., 2023) and MultiArith (Roy and Roth, 2015). As in AgentCoMa, the questions in these benchmarks contain exactly two reasoning steps; we extract them manually as individual sub-questions, each associated with its (intermediate) ground-truth answer (further details are in Appendix J). Note that the math steps in MultiArith and AgentCoMa are of similar difficulty: both require a single small-number arithmetic operation. In Figure 4 we compare performance between the three benchmarks, on the individual steps and their composition (see Table 11 for numerical results).

We observe that LLM performance on MultiArith is near-perfect and approximately uniform across steps and composition (<1% avg. compositionality gap). On Bamboogle, LLMs achieve 69% mean accuracy on *each* individual step, but can independently solve *both* in only 53% of cases on average. As the mean compositional accuracy is 52%, the gap is negligible[3]. Crucially, we find that over 80% of compositional failures in Bamboogle are due to the LLMs not being able to answer correctly either one or both of the underlying steps (see Figure 5(b)). In contrast, AgentCoMa specifically unveils *compositional* failures, where models giving incorrect answers usually solve the individual steps correctly, as observed in Section 5.

[3]Note that a minority of the LLMs in Figure 4 is more accurate on the composition than on both steps independently.

## 6 Analysis of the Compositionality Gap

### 6.1 Effect of Additional Context

Prior work has shown that adding more context to LLM inputs can sometimes hurt their performance (Wu et al., 2024). Therefore, we investigate whether the additional context that derives from merging multiple reasoning steps contributes to the compositionality gap in AgentCoMa. It should be noted that second-step (math) sub-questions are already approximately equal in length to the compositional ones (see average sequence lengths in Appendix K), as they include all the context (and the intermediate answer) of the first (commonsense) step. For analysis purposes, we thus also lengthen the commonsense sub-questions to match the compositional ones, by adding the numerical information that would be necessary to solve the mathematical step, but is not needed for the commonsense part (see examples in Appendix K). We run all LLMs on these modified commonsense sub-questions to quantify the confounding effect of the extra context. This is a particularly challenging baseline, as the additional context here is also *irrelevant*, which has been shown to reduce LLM performance dramatically (Shi et al., 2023). As shown in Figure 6, most LLMs see a performance decrease (greater for smaller models) on these sub-questions compared to the original commonsense ones. Nevertheless, this drop is moderate compared to the substantial performance degradation we observe in the compositional task. Therefore, the compositionality gap in AgentCoMa cannot be mainly attributed to the longer context.

| Question type | Ratio | Var | Std |
|---|---|---|---|
| Commonsense | 71.49 | 0.0004 | 0.0209 |
| Math | 72.20 | 0.0008 | 0.0285 |
| Composition | 70.75 | 0.0009 | 0.0298 |

Table 3: Lookback attention ratios for AgentCoMa, with variance and standard deviation.

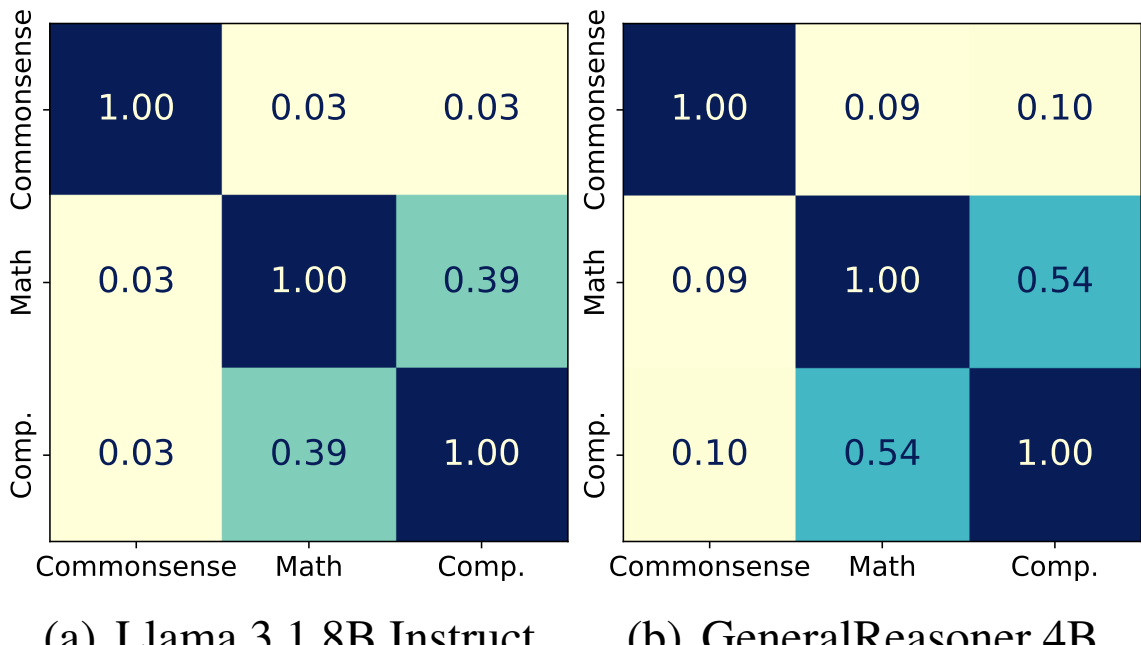

(a) Llama 3.1 8B Instruct (b) GeneralReasoner 4B

Figure 8: Neuron overlap rates between tasks in AgentCoMa for (a) Llama 3.1 8B Instruct and (b) GeneralReasoner 4B. While the compositional task has substantial neuron overlap with the math reasoning step (39% and 54%, respectively), it has little overlap with the commonsense step (3% and 10%, respectively).

### 6.2 Similarity to the Training Data

As language models are known to perform less well on unseen patterns (Yang et al., 2023), another plausible hypothesis is that poor performance on AgentCoMa may be due to the absence of similar mixed-type compositional reasoning tasks in the models' training data. To investigate this, we perform membership inference analysis (MIA) (Shokri et al., 2017) on both the compositional questions and the individual steps.

We use Min-K%++ (Zhang et al., 2025a), which achieves state-of-the-art performance in training data detection at the time of writing. Min-K%++ evaluates how "surprised" a model is by a given text: it selects the lowest-scoring K% of next-token predictions and averages their normalised scores, placing emphasis on the least predictable tokens. Importantly, this does not test whether a sequence appeared verbatim in the training data; rather, the scores are influenced by recurring patterns or partial overlaps, even when the full sequence is novel. Although Zhang et al. (2025a) apply Min-K%++ with a binary threshold for detection, the metric can also be used comparatively as a more robust alternative to perplexity: a statistically higher score for one corpus over another indicates closer alignment with patterns the model likely encountered during training. Figure 7 shows the results of this analysis, averaged across instruction-tuned models and reasoning models (both SFT and RL). For both groups, the Min-K%++ scores are lower for the compositional questions than for the individual steps, with a slightly greater discrepancy for reasoning models. This suggests that mixed-type reasoning tasks are relatively rare in the training data, which may contribute to the weaker performance on AgentCoMa.

### 6.3 Attention Maps Analysis

We observe that the incorrect compositional outputs generated by the LLMs, albeit fluent and plausible at first glance, contain reasoning steps that are inconsistent with the given context upon closer inspection (we show an example of these outputs in Appendix L.2). We thus hypothesise that the LLMs suffer from *contextual hallucination* (Chuang et al., 2024) when presented with the compositional questions in AgentCoMa, i.e., they fail to correctly process and utilise the context. We perform lookback attention analysis (Chuang et al., 2024) and compute a ratio that represents how much weight the LLM assigns to the context while generating a response (due to the computationally intensive nature of this analysis, we perform it with Llama 3.1 8B Instruct). The results are shown in Table 3. Indeed, we find that the lookback attention ratio is lower for the compositional questions than for the individual steps (all pairwise differences are statistically significant; see Appendix M). This indicates that less attention is paid to the compositional inputs, leading to higher contextual hallucination in the mixed-type reasoning task.

### 6.4 Neuron Pattern Analysis

So far we have shown that LLMs can largely solve the individual sub-questions in AgentCoMa, which are more aligned with their training data, yet they suffer from contextual hallucination when presented with their composition. A potential explanation is that at inference the models fail to activate all the neurons needed for mixed-type reasoning, instead relying on neural circuits appropriate for only one reasoning type—as they may be conditioned to do during training. To test this hypothesis, we use Query-Relevant Neuron Cluster Attribution (QRNCA) (Chen et al., 2025a) to identify which neurons influence the model's in-ternal representation of a given (question, answer) pair. Specifically, QRNCA measures the contribution of individual neurons to the model's encoding of the answer given the question, using the ground truth answer rather than the model's own prediction. Chen et al. (2025a) demonstrate that queries from different domains tend to activate distinct neuron regions, from which it follows that a query combining domains should activate both corresponding regions. Therefore, if LLMs are able to combine the appropriate neurons for mixed-type reasoning, we would expect the neuron patterns of the individual steps to overlap substantially with that of the compositional task.

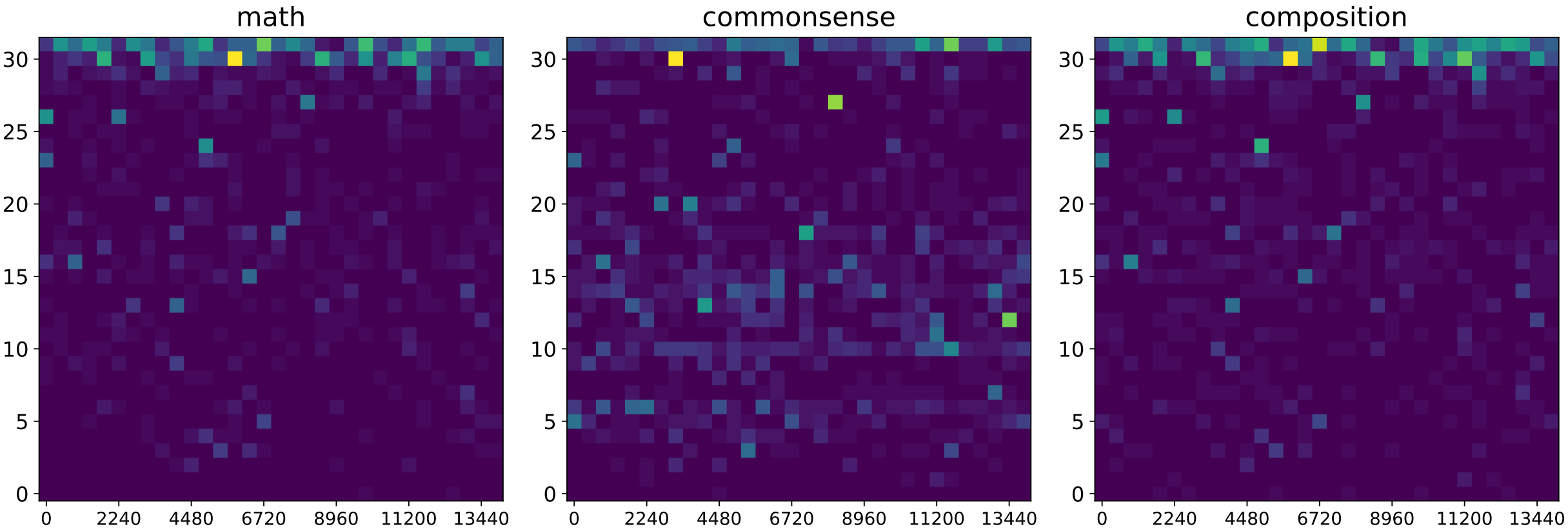

Figure 9: Location patterns showing the active neurons in the individual reasoning steps and the compositional questions, for Llama 3.1 8B Instruct. The compositional pattern resembles the math one, however, the commonsense neurons remain largely inactive.

In Figure 8, we show the average neuron overlap rates between the compositional questions and the individual steps for Llama 3.1 8B Instruct (Figure 8(a)) and GeneralReasoner 4B (Figure 8(b)) (due to the computational effort required to perform QRNCA, we limit this analysis to two models, taking care in choosing models that differ in both size and training paradigm). We observe that both models leverage largely distinct neurons when solving the math and the commonsense steps (only 3% overlap for Llama 3.1 8B Instruct and 9% for GeneralReasoner 4B). Notably, we also find that the neurons leveraged to solve the compositional task overlap substantially with those active during the math reasoning step (39% for Llama 3.1 8B Instruct and 54% for GeneralReasoner 4B), but have only a negligible overlap with those active in the commonsense step (3% for Llama 3.1 8B Instruct and 10% for GeneralReasoner 4B).

In Figure 9, we show the locations of Llama 3.1 8B Instruct neurons relevant to each sub-question in AgentCoMa (math, commonsense), as well as the neurons active during compositional queries. We observe that the compositional neuron map closely resembles the math one, confirming the relatively high overlap rate illustrated in Figure 8. However, neurons relevant to commonsense sub-questions appear largely inactive during the compositional task, consistent with the negligible overlap rate in Figure 8. The neuron activation heatmaps for GeneralReasoner 4B follow a similar pattern. Overall, the LLM fails to leverage the appropriate neurons for both reasoning types when attempting the compositional questions, only activating neurons relevant to one type of reasoning (math).

## 7 Conclusion

We introduce AgentCoMa, a new benchmark that requires the composition of commonsense and mathematical reasoning in real-world scenarios. Through a large-scale evaluation of 61 models, we show that LLMs tend to perform poorly on its questions, despite achieving high accuracy on their component steps in isolation. This performance gap is substantially greater than in compositional benchmarks requiring only one type of reasoning. Our analysis reveals that tasks mixing commonsense and math reasoning are relatively rare in LLM training data. This unfamiliarity appears to lead the models to activate neurons relevant to only one reasoning type when they attempt to solve AgentCoMa, as a result of patterns likely reinforced during training. Our work supports future advances in training methods for agentic reasoning, while providing a rigorous test bed for their evaluation.

## Limitations

AgentCoMa is intentionally designed as a controlled experiment to isolate the effect of combining two distinct types of reasoning. Its structured nature enables us to derive simplified sub-questions that each require only one reasoning type, allowing us to benchmark performance on the compositional queries directly against performance on the individual steps. Real-world reasoning tasks are, of course, often more complex and less neatly separable. However, without a controlled framework like AgentCoMa, it would be difficult to determine whether differences in model behaviour stem from the challenge of combining reasoning types or from other confounding factors present in naturalistic data. We view this deliberately simplified setting as a necessary foundation for studying mixed-type compositional reasoning, one that can subsequently inform the analysis of longer, more realistic tasks.

In Appendix N, we present a preliminary investigation into how the order of reasoning steps affects performance. This experiment is small in scale, as constructing a full reversed-order dataset would require substantial expert annotation effort and falls outside the current scope. We hope that future work will expand on this direction, and potentially incorporate additional reasoning types relevant to agentic settings.

## Ethical Considerations

We have carefully verified that the software, model checkpoints and existing datasets utilised in this work are permitted for access, distribution and, where relevant, modification. Our use and purpose comply with those terms. All samples in AgentCoMa are written by human experts without the aid of automated tools, ensuring originality and independence from existing licensed sources. We have verified that the dataset contains no offensive or harmful content; all samples are grounded in neutral, real-world scenarios.

## Acknowledgments

The computations described in this research were performed using the Baskerville Tier 2 HPC service. Baskerville was funded by the EPSRC and UKRI through the World Class Labs scheme (EP/T022221/1) and the Digital Research Infrastructure programme (EP/W032244/1) and is operated by Advanced Research Computing at the University of Birmingham. The authors would like to thank the Alan Turing Institute for facilitating access to this HPC facility. Ana Brassard's effort was supported by a RIKEN Incentive Research Project FY2024. Hossein A. Rahmani's effort was supported by the Engineering and Physical Sciences Research Council (EP/S021566/1). Marek Rei is supported in part by Gen AI – the AI Hub for Generative Models, funded by EPSRC (EP/Y028805/1).

## A Benchmark Results

In Tables 4 and 5 we show the accuracies on the test and development sets, respectively, of AgentCoMa. Accuracies are shown for 61 LLMs, grouped into instruction-tuned models, reasoning models optimised mainly via supervised fine-tuning (SFT) and reasoning models optimised mainly via online reinforcement learning (RL). Within each group, we order the LLMs by increasing total number of parameters.

## B Preliminary Experiment

We experiment with prompting GPT-4o to automate the creation of new questions. These preliminary experiments are unsuccessful. Even when it is provided with high-quality question-answer pairs as exemplars, the model produces samples that do not require a choice based on commonsense, effectively returning math-only questions. Below we show an input prompt with the corresponding model output.

***Please give me questions similar to the one below. The questions need to be about housework, such as cleaning or other house chores, and need to require common sense knowledge (like in the question below the response needs to know that carpets do not get mopped), and then they need to do a math calculation (multiplication) with the selected numbers, like in the question below. Give me five similar questions and corresponding answers. Do not make them too similar, not about tiles and carpet, and not necessarily about cleaning floors. They have to require some common knowledge of the characteristics of the objects you discuss.***

***Question: Youre deep cleaning your kitchen and bedroom, which includes mopping floors where possible. The kitchen floor is tiled, 3.5 m wide and 4 m long, while the bedroom floor has a carpet and is 4.5 m by 5 m. How many square meters of floor do you mop?***

***Answer: You cant mop carpet, only tiles, so you will only mop the kitchen. So you mop 3.5*** $\times$***4 = 14 m***$^2$***. So the final answer is: 14***

*Great! Here's a fresh set of five word problems that:*

*Are based on housework or chores*

*Require commonsense knowledge (like what materials can be ironed, washed, or vacuumed, etc.)*

*Involve multiplication to calculate the final result*

*Arent just about floors or carpet*

*Lets get into it:*

*1. Question:*

*You're ironing clothes from two laundry baskets. One basket has cotton shirts, and the other has wool sweaters. Each cotton shirt takes 3 minutes to iron, and there are 6 of them. The wool sweaters are delicate and shouldnt be ironed. How many minutes will you spend ironing?*

*Answer: Wool sweaters shouldnt be ironed, so only the cotton shirts are ironed. You iron 6 shirts* $\times$*3 minutes = 18 minutes. Final answer: 18*

*2. Question:*

*Youre organizing your living room and vacuuming the furniture. You have 3 fabric armchairs and 2 leather armchairs. Each fabric armchair takes 5 minutes to vacuum. Leather armchairs just need wiping, not vacuuming. How many minutes will you spend vacuuming the armchairs?*

*Answer: Only fabric armchairs are vacuumed. 3 fabric chairs* $\times$*5 minutes = 15 minutes. Final answer: 15*

*3. Question:*

*Youre cleaning your windows. You have 4 large windows in the living room and 3 skylights in the attic. Each large window takes 6 minutes to clean. Skylights are too high to reach without a ladder, which you dont have today. How many minutes will you spend cleaning windows?*

*Answer: Only the 4 large windows can be cleaned. 4* $\times$*6 minutes = 24 minutes. Final answer: 24*

*4. Question:*

*You're washing curtains from the house. The 5 curtains in the dining room are machine-washable. The 3 curtains in the bedroom are dry-clean only and cant go in your washing machine. Each machine-washable curtain takes 40 minutes to wash. How many minutes of washing will you do?*

*Answer: Only the 5 dining room curtains are washable at home. 5* $\times$*40 = 200 minutes. Final answer: 200*

*5. Question:*

*You're polishing wooden furniture in your home. The living room has 2 wooden tables and 4 plastic chairs. Each wooden table takes 12 minutes to polish. Plastic chairs dont get polished. How many minutes will you spend polishing furniture?*

*Answer: Only wooden furniture is polished. 2* $\times$*12 = 24 minutes. Final answer: 24*

*Let me know if you'd like more problems that use division, or combine multiple operations with reasoning!*

| Model Name | Model Type | First step | Second step | Both correct | Composition |
|---|---|---|---|---|---|
| Qwen3 1.7B | | 47.8 | 62.2 | 30.6 | 12.2 |
| Llama3.2 3B Instruct | | 67.2 | 66.7 | 46.7 | 25.6 |
| Qwen2.5 3B Instruct | | 72.8 | 78.3 | 57.8 | 33.9 |
| Phi3.5 Mini 3.8B Instruct | | 83.9 | 86.7 | 73.3 | 42.2 |
| Phi4 Mini 3.8B Instruct | | 77.8 | 86.7 | 66.1 | 35.6 |
| Qwen3 4B | | 76.1 | 88.9 | 67.8 | 35.6 |
| Command R7B | | 70.6 | 80.0 | 58.3 | 35.0 |
| OLMo2 7B Instruct | | 74.4 | 66.7 | 46.7 | 21.1 |
| Mistral 7B Instruct | | 84.4 | 75.6 | 63.9 | 31.1 |
| Qwen2.5 7B Instruct | | 82.8 | 87.8 | 73.9 | 51.7 |
| Ministral 8B Instruct | | 79.4 | 86.1 | 68.9 | 37.2 |
| Llama3.1 8B Instruct | | 81.1 | 83.9 | 68.3 | 33.3 |
| Qwen3 8B | | 86.1 | 92.8 | 81.1 | 50.0 |
| Mistral NeMo 12.2B Instruct | *Instruction-tuned* | 82.2 | 88.9 | 75.0 | 45.0 |
| OLMo2 13B Instruct | | 82.2 | 62.2 | 52.8 | 23.9 |
| Qwen2.5 14B Instruct | | 93.3 | 67.8 | 62.2 | 42.8 |
| Qwen3 14B | | 93.3 | 94.4 | 88.9 | 60.6 |
| Mistral Small 22B Instruct | | 90.6 | 90.6 | 82.8 | 56.7 |
| Qwen3 30B A3B | | 91.1 | 96.1 | 87.8 | 52.2 |
| Qwen3 32B | | 95.6 | 95.0 | 91.7 | 56.7 |
| OLMo2 32B Instruct | | 89.4 | 86.1 | 77.2 | 36.1 |
| Yi1.5 34B Chat | | 87.2 | 87.8 | 76.1 | 41.7 |
| Phi3.5 MoE 42B | | 96.1 | 91.1 | 87.2 | 61.7 |
| Llama3.3 70B Instruct | | 93.9 | 96.1 | 90.0 | 73.3 |
| Tülu3 70B | | 94.4 | 93.9 | 88.3 | 66.7 |
| Hermes3 70B | | 91.1 | 95.6 | 87.8 | 66.1 |
| Qwen2.5 72B Instruct | | 96.1 | 93.9 | 90.0 | 68.3 |
| AceInstruct 72B | | 90.6 | 92.2 | 84.4 | 56.1 |
| Command R+ 104B | | 84.4 | 83.9 | 72.8 | 46.1 |
| Mixtral MoE 141B | | 95.6 | 95.0 | 90.6 | 66.1 |
| SmallThinker 3B | | 63.9 | 72.2 | 46.7 | 23.9 |
| Art 3B | | 68.3 | 83.9 | 57.8 | 33.9 |
| Mathstral 7B | | 79.4 | 90.6 | 72.2 | 36.1 |
| DeepSeekR1 Distill 7B | | 63.9 | 75.6 | 50.0 | 23.9 |
| DeepSeekR1 Distill 8B | | 76.1 | 76.7 | 59.4 | 27.2 |
| DeepSeekR1 8B | *Reasoning (SFT)* | 90.6 | 80.6 | 72.2 | 34.4 |
| DeepSeekR1 Distill 14B | | 90.6 | 93.9 | 85.6 | 61.7 |
| Phi4 Reasoning 14.7B | | 95.6 | 95.0 | 91.7 | 62.2 |
| DeepSeekR1 Distill 32B | | 94.4 | 95.6 | 90.0 | 60.0 |
| OpenThinker2 32B | | 62.2 | 71.1 | 43.9 | 15.0 |
| Sky-T1 32B | | 91.7 | 96.7 | 88.3 | 65.0 |
| DeepSeekR1 Distill 70B | | 96.1 | 91.7 | 88.3 | 71.7 |
| Reflection-Llama 70B | | 91.1 | 90.6 | 82.2 | 65.6 |
| Command A 111B | | 96.7 | 96.1 | 92.8 | 71.1 |
| SimpleRL 1.5B | | 45.0 | 59.4 | 27.8 | 12.8 |
| EXAONE Deep 2.4B | | 56.1 | 58.3 | 36.1 | 23.3 |
| GeneralReasoner 4B | | 82.2 | 87.8 | 73.9 | 36.7 |
| DeepSeek Math 7B Instruct | | 45.0 | 71.1 | 31.1 | 12.8 |
| GeneralReasoner 7B | | 83.3 | 78.3 | 65.6 | 35.6 |
| SimpleRL 7B | | 84.4 | 78.3 | 66.7 | 36.1 |
| SimpleRL Math 7B | | 53.9 | 77.8 | 50.0 | 21.7 |
| Marco-o1 7.6B | | 80.0 | 86.1 | 70.0 | 41.7 |
| SimpleRL 8B | *Reasoning (RL)* | 68.9 | 79.4 | 56.7 | 25.0 |
| GeneralReasoner 14B | | 91.1 | 88.3 | 80.0 | 46.1 |
| SimpleRL 14B | | 89.4 | 83.9 | 76.7 | 58.3 |
| UniReason 14B RL | | 89.4 | 88.9 | 78.9 | 51.1 |
| Phi4 Reasoning Plus 14.7B | | 95.0 | 96.1 | 91.7 | 62.2 |
| SimpleRL 24B | | 87.2 | 91.7 | 78.9 | 61.7 |
| EXAONE Deep 32B | | 92.2 | 78.9 | 73.3 | 36.7 |
| SimpleRL 32B | | 95.6 | 97.8 | 93.9 | 66.7 |
| QVQ 72B | | 93.9 | 92.8 | 87.8 | 56.7 |

Table 4: LLM accuracies on the test set of AgentCoMa. We perform greedy sampling with all models.

| Model Name | Model Type | First step | Second step | Both correct | Composition |
|---|---|---|---|---|---|
| Qwen3 1.7B | | 48.8 | 70.0 | 32.5 | 10.0 |
| Llama3.2 3B Instruct | | 62.5 | 71.2 | 43.8 | 21.2 |
| Qwen2.5 3B Instruct | | 70.0 | 72.5 | 50.0 | 32.5 |
| Phi3.5 Mini 3.8B Instruct | | 87.5 | 83.8 | 72.5 | 33.8 |
| Phi4 Mini 3.8B Instruct | | 61.2 | 85.0 | 53.8 | 27.5 |
| Qwen3 4B | | 77.5 | 82.5 | 67.5 | 27.5 |
| Command R7B | | 66.2 | 85.0 | 57.5 | 35.0 |
| OLMo2 7B Instruct | | 70.0 | 67.5 | 48.8 | 12.5 |
| Mistral 7B Instruct | | 80.0 | 73.8 | 60.0 | 33.8 |
| Qwen2.5 7B Instruct | | 78.8 | 81.2 | 66.3 | 42.5 |
| Ministral 8B Instruct | | 78.8 | 83.8 | 66.3 | 35.0 |
| Llama3.1 8B Instruct | | 72.5 | 86.2 | 63.8 | 36.2 |
| Qwen3 8B | | 82.5 | 91.2 | 78.8 | 45.0 |
| Mistral NeMo 12.2B Instruct | *Instruction-tuned* | 83.8 | 86.2 | 72.5 | 51.2 |
| OLMo2 13B Instruct | | 67.5 | 60.0 | 38.8 | 16.2 |
| Qwen2.5 14B Instruct | | 87.5 | 68.8 | 61.3 | 48.8 |
| Qwen3 14B | | 90.0 | 95.0 | 87.5 | 51.2 |
| Mistral Small 22B Instruct | | 88.8 | 87.5 | 78.8 | 48.8 |
| Qwen3 30B A3B | | 93.8 | 93.8 | 87.5 | 46.2 |
| Qwen3 32B | | 88.8 | 93.8 | 82.5 | 48.8 |
| OLMo2 32B Instruct | | 88.8 | 83.8 | 73.8 | 33.8 |
| Yi1.5 34B Chat | | 80.0 | 91.2 | 72.5 | 47.5 |
| Phi3.5 MoE 42B | | 91.2 | 90.0 | 82.5 | 48.8 |
| Llama3.3 70B Instruct | | 93.8 | 96.2 | 91.3 | 76.2 |
| Tülu3 70B | | 92.5 | 95.0 | 87.5 | 61.2 |
| Hermes3 70B | | 92.5 | 93.8 | 86.3 | 66.2 |
| Qwen2.5 72B Instruct | | 93.8 | 91.2 | 86.3 | 56.2 |
| AceInstruct 72B | | 87.5 | 92.5 | 81.3 | 46.2 |
| Command R+ 104B | | 83.8 | 87.5 | 73.8 | 41.2 |
| Mixtral MoE 141B | | 93.8 | 91.2 | 86.3 | 63.8 |
| SmallThinker 3B | | 68.8 | 77.5 | 53.8 | 23.8 |
| Art 3B | | 67.5 | 83.8 | 57.5 | 31.2 |
| Mathstral 7B | | 76.2 | 85.0 | 65.0 | 31.2 |
| DeepSeekR1 Distill 7B | | 58.8 | 75.0 | 48.8 | 22.5 |
| DeepSeekR1 Distill 8B | | 60.0 | 70.0 | 45.0 | 25.0 |
| DeepSeekR1 8B | *Reasoning (SFT)* | 87.5 | 85.0 | 78.8 | 32.5 |
| DeepSeekR1 Distill 14B | | 78.8 | 90.0 | 72.5 | 61.2 |
| Phi4 Reasoning 14.7B | | 93.8 | 96.2 | 90.0 | 60.0 |
| DeepSeekR1 Distill 32B | | 90.0 | 92.5 | 85.0 | 60.0 |
| OpenThinker2 32B | | 57.5 | 71.2 | 48.8 | 15.0 |
| Sky-T1 32B | | 95.0 | 90.0 | 85.0 | 61.2 |
| DeepSeekR1 Distill 70B | | 92.5 | 95.0 | 87.5 | 71.2 |
| Reflection-Llama 70B | | 90.0 | 93.8 | 85.0 | 62.5 |
| Command A 111B | | 93.8 | 93.8 | 87.5 | 60.0 |
| SimpleRL 1.5B | | 40.0 | 62.5 | 23.8 | 8.8 |
| EXAONE Deep 2.4B | | 53.8 | 60.0 | 31.3 | 17.5 |
| GeneralReasoner 4B | | 77.5 | 88.8 | 68.8 | 38.8 |
| DeepSeek Math 7B Instruct | | 33.8 | 75.0 | 22.5 | 6.3 |
| GeneralReasoner 7B | | 80.0 | 78.8 | 61.3 | 22.5 |
| SimpleRL 7B | | 80.0 | 77.5 | 58.8 | 33.8 |
| SimpleRL Math 7B | | 46.2 | 75.0 | 37.5 | 16.2 |
| Marco-o1 7.6B | | 77.5 | 85.0 | 66.3 | 37.5 |
| SimpleRL 8B | *Reasoning (RL)* | 62.5 | 73.8 | 46.3 | 22.5 |
| GeneralReasoner 14B | | 90.0 | 87.5 | 78.8 | 50.0 |
| SimpleRL 14B | | 90.0 | 90.0 | 80.0 | 51.2 |
| UniReason 14B RL | | 82.5 | 87.5 | 71.3 | 46.2 |
| Phi4 Reasoning Plus 14.7B | | 91.2 | 97.5 | 90.0 | 56.2 |
| SimpleRL 24B | | 88.8 | 92.5 | 81.3 | 62.5 |
| EXAONE Deep 32B | | 92.5 | 68.8 | 63.8 | 28.8 |
| SimpleRL 32B | | 88.8 | 95.0 | 83.8 | 58.8 |
| QVQ 72B | | 88.8 | 95.0 | 83.8 | 56.2 |

Table 5: LLM accuracies on the development set of AgentCoMa. We perform greedy sampling with all models.

| Model Name | Model Type | HuggingFace Model Identifier |
|---|---|---|
| Qwen3 1.7B | | `rd211/Qwen3-1.7B-Instruct` |
| Llama3.2 3B Instruct | | `meta-llama/Llama-3.2-3B-Instruct` |
| Qwen2.5 3B Instruct | | `Qwen/Qwen2.5-3B-Instruct` |
| Phi3.5 Mini 3.8B Instruct | | `microsoft/Phi-3.5-mini-instruct` |
| Phi4 Mini 3.8B Instruct | | `microsoft/Phi-4-mini-instruct` |
| Qwen3 4B | | `Qwen/Qwen3-4B` |
| Command R7B | | `CohereLabs/c4ai-command-r7b-12-2024` |
| OLMo2 7B Instruct | | `allenai/OLMo-2-1124-7B-Instruct` |
| Mistral 7B Instruct | | `mistralai/Mistral-7B-Instruct-v0.3` |
| Qwen2.5 7B Instruct | | `Qwen/Qwen2.5-7B-Instruct` |
| Ministral 8B Instruct | | `mistralai/Ministral-8B-Instruct-2410` |
| Llama3.1 8B Instruct | | `meta-llama/Llama-3.1-8B-Instruct` |
| Qwen3 8B | | `Qwen/Qwen3-8B` |
| Mistral NeMo 12.2B Instruct | *Instruction-tuned* | `mistralai/Mistral-Nemo-Instruct-2407` |
| OLMo2 13B Instruct | | `allenai/OLMo-2-1124-13B-Instruct` |
| Qwen2.5 14B Instruct | | `Qwen/Qwen2.5-14B-Instruct` |
| Qwen3 14B | | `Qwen/Qwen3-14B` |
| Mistral Small 22B Instruct | | `mistralai/Mistral-Small-Instruct-2409` |
| Qwen3 30B A3B | | `Qwen/Qwen3-30B-A3B` |
| Qwen3 32B | | `Qwen/Qwen3-32B` |
| OLMo2 32B Instruct | | `allenai/OLMo-2-0325-32B-Instruct` |
| Yi1.5 34B Chat | | `01-ai/Yi-1.5-34B-Chat` |
| Phi3.5 MoE 42B | | `microsoft/Phi-3.5-MoE-instruct` |
| Llama3.3 70B Instruct | | `meta-llama/Llama-3.3-70B-Instruct` |
| Tülu3 70B | | `allenai/Llama-3.1-Tulu-3-70B` |
| Hermes3 70B | | `NousResearch/Hermes-3-Llama-3.1-70B` |
| Qwen2.5 72B Instruct | | `Qwen/Qwen2.5-72B-Instruct` |
| AceInstruct 72B | | `nvidia/AceInstruct-72B` |
| Command R+ 104B | | `CohereLabs/c4ai-command-r-plus` |
| Mixtral MoE 141B | | `mistralai/Mixtral-8x22B-Instruct-v0.1` |
| SmallThinker 3B | | `PowerInfer/SmallThinker-3B-Preview` |
| Art 3B | | `AGI-0/Art-v0-3B` |
| Mathstral 7B | | `mistralai/Mathstral-7B-v0.1` |
| DeepSeekR1 Distill 7B | | `deepseek-ai/DeepSeek-R1-Distill-Qwen-7B` |
| DeepSeekR1 Distill 8B | | `deepseek-ai/DeepSeek-R1-Distill-Llama-8B` |
| DeepSeekR1 8B | *Reasoning (SFT)* | `deepseek-ai/DeepSeek-R1-0528-Qwen3-8B` |
| DeepSeekR1 Distill 14B | | `deepseek-ai/DeepSeek-R1-Distill-Qwen-14B` |
| Phi4 Reasoning 14.7B | | `microsoft/Phi-4-reasoning` |
| DeepSeekR1 Distill 32B | | `deepseek-ai/DeepSeek-R1-Distill-Qwen-32B` |
| OpenThinker2 32B | | `open-thoughts/OpenThinker2-32B` |
| Sky-T1 32B | | `NovaSky-AI/Sky-T1-32B-Flash` |
| DeepSeekR1 Distill 70B | | *deepseek-ai/DeepSeek-R1-Distill-Llama-70B* |
| Reflection-Llama 70B | | `mattshumer/Reflection-Llama-3.1-70B` |
| Command A 111B | | `CohereLabs/c4ai-command-a-03-2025` |
| SimpleRL 1.5B | | `hkust-nlp/Qwen-2.5-1.5B-SimpleRL-Zoo` |
| EXAONE Deep 2.4B | | `LGAI-EXAONE/EXAONE-Deep-2.4B` |
| GeneralReasoner 4B | | `TIGER-Lab/General-Reasoner-Qwen3-4B` |
| DeepSeek Math 7B Instruct | | `deepseek-ai/deepseek-math-7b-instruct` |
| GeneralReasoner 7B | | `TIGER-Lab/General-Reasoner-Qwen2.5-7B` |
| SimpleRL 7B | | `hkust-nlp/Qwen-2.5-7B-SimpleRL-Zoo` |
| SimpleRL Math 7B | | `hkust-nlp/DeepSeek-Math-7B-SimpleRL-Zoo` |
| Marco-o1 7.6B | | `AIDC-AI/Marco-o1` |
| SimpleRL 8B | *Reasoning (RL)* | `hkust-nlp/Llama-3.1-8B-SimpleRL-Zoo` |
| GeneralReasoner 14B | | `TIGER-Lab/General-Reasoner-Qwen3-14B` |
| SimpleRL 14B | | `hkust-nlp/Qwen-2.5-14B-SimpleRL-Zoo` |
| UniReason 14B RL | | `ReasoningTransferability/UniReason-Qwen3-14B-RL` |
| Phi4 Reasoning Plus 14.7B | | `microsoft/Phi-4-reasoning-plus` |
| SimpleRL 24B | | `hkust-nlp/Mistral-Small-24B-SimpleRL-Zoo` |
| EXAONE Deep 32B | | `LGAI-EXAONE/EXAONE-Deep-32B` |
| SimpleRL 32B | | `hkust-nlp/Qwen-2.5-32B-SimpleRL-Zoo` |
| QVQ 72B | | `Qwen/QVQ-72B-Preview` |

Table 6: HuggingFace model identifiers of all LLMs benchmarked on AgentCoMa.

## C Question Decomposition

All inference-time experiments discussed in Sections 5 and 5.1 use CoT prompting, which is known to aid LLMs in compositional tasks (Li et al., 2023, 2024). On the other hand, other prompting strategies have been proposed for this purpose, such as *self-ask* (Press et al., 2023), which encourages the model to decompose a complex task into manageable sub-tasks and has been shown to substantially reduce the compositionality gap on a variety of benchmarks, including Bamboogle. In Table 7, we show the results of running *self-ask* on AgentCoMa, using the same 16 LLMs as in Table 2. Even with *self-ask*, the compositionality gap remains nearly as high (27% avg.) as the one observed using CoT (29% avg.). Appendix E.2 illustrates the Self-Ask prompts.

## D Model IDs and Hyperparameters

Table 6 lists the HuggingFace model identifiers for all 61 LLMs benchmarked in this paper. We show the inference hyperparameters in Table 8. We run all inference on a GPU node with four Nvidia A100s, using the `vLLM` library, and apply greedy decoding. We also allow for a maximum generation budget of 2048 tokens, which is large considering that the answers to AgentCoMa questions only require two reasoning steps to solve. This is to prevent the LLM responses from being cut off, particularly for reasoning models.

| Model Name | Model Type | Compositional accuracy (self-ask) |
|---|---|---|
| Phi4 Mini 3.8B | *Instruction-tuned* | 35.0 |
| Llama3.1 8B IT | | 44.4 |
| Qwen3 14B | | 60.6 |
| Phi3.5 MoE 42B | | 56.1 |
| Llama3.3 70B IT | | 77.2 |
| Mixtral MoE 141B | | 66.7 |
| Art 3B | *Reasoning (SFT)* | 20.0 |
| DeepSeekR1 8B | | 40.6 |
| Phi4 Reasoning 14.7B | | 67.8 |
| DeepSeekR1 32B | | 68.3 |
| Reflection-Llama 70B | | 69.7 |
| GeneralReasoner 4B | *Reasoning (RL)* | 36.7 |
| SimpleRL 8B | | 31.7 |
| GeneralReasoner 14B | | 53.9 |
| SimpleRL 32B | | 57.2 |
| QVQ 72B | | 60.6 |

Table 7: Accuracy scores of eliciting answers via *self-ask* on the compositional questions in AgentCoMa, compared to the individual steps. We perform greedy sampling with all LLMs.

| Hyperparameter | Value |
|---|---|
| Temperature | 0 |
| Top-$k$ | 1 |
| Max model length | 4096 |
| Max tokens | 2048 |

Table 8: Inference hyperparameters for all models.

## E Prompts

### E.1 Inference Prompts (CoT)

We use two-shot CoT prompts to run all inference on AgentCoMa, Bamboogle and MultiArith. For AgentCoMa, we write two ad-hoc examples to add to the prompt. These do not appear in either the test set or the development set. Since Bamboogle also does not have a training set, we use two questions created by Press et al. (2023) specifically for running inference on this benchmark, with corresponding CoT-style answers. For MultiArith, we utilise two examples from the training set. We show all prompts below.

**AgentCoMa compositional CoT prompt.**

```
Answer the questions below step by step.
Always conclude with the sentence 'So the
final answer is:' followed by the final
numerical answer.

Question: You have a buffet plate
that is 1.3 cm thick, a saucer that is 0.7
cm thick, a dinner plate that is 1.65 cm
thick, and a dessert plate that is 0.93 cm
thick.
You want to stack them by decreasing
diameter, the widest at the bottom. What
will be the combined height of the two
plates at the top of the stack, in
centimeters?

Answer: For the plates to be stacked
by decreasing diameter, the stacking order
(from bottom to top) is:
Buffet plate (1.3 cm thick), dinner plate
(1.65 cm thick), dessert plate (0.93 cm
thick), saucer (0.7 cm thick).
Hence the two plates at the top of the
stack are the dessert plate and the saucer,
and their combined height is 0.93 + 0.7 =
1.63 cm.
```

So the final answer is: 1.63

Question: Kuala Lumpur has 4500 restaurants, Albuquerque has 653 restaurants, Dubai has 13,000 restaurants. How many restaurants do the Asian cities have, combined?

Answer: Kuala Lumpur and Dubai are in Asia, whereas Albuquerque is in America. So the Asian cities are Kuala Lumpur and Dubai.
Kuala Lumpur has 4500 restaurants and Dubai has 13,000. So combined they have 4500+13000=17500 restaurants.
So the final answer is: 17500

Question: {question}

Answer:

### AgentCoMa 1st step CoT prompt.

Answer the questions below step by step. Always conclude with the sentence ’So the final answer is:’ followed by the final answer.

Question: You have a buffet plate, a saucer, a dinner plate, and a dessert plate. You want to stack them by decreasing diameter, the widest at the bottom. Which two plates are at the top of the stack?

Answer: For the plates to be stacked by decreasing diameter, the stacking order (from bottom to top) is:
Buffet plate, dinner plate, dessert plate, saucer.
Hence the two plates at the top of the stack are the dessert plate and the saucer.
So the final answer is: dessert plate and saucer

Question: Kuala Lumpur has 4500 restaurants, Albuquerque has 653 restaurants, Dubai has 13,000 restaurants. Which of these are Asian cities?

Answer: Kuala Lumpur and Dubai are in Asia, whereas Albuquerque is in America. So the Asian cities are Kuala Lumpur and Dubai.
So the final answer is: Kuala Lumpur and Dubai

Question: {question}

Answer:

### AgentCoMa 2nd step CoT prompt.

Answer the questions below step by step. Always conclude with the sentence ’So the final answer is:’ followed by the final numerical answer.

Question: You have a buffet plate that is 1.3 cm thick, a saucer that is 0.7 cm thick, a dinner plate that is 1.65 cm thick, and a dessert plate that is 0.93 cm thick. You want to stack them by decreasing diameter, the widest at the bottom. What will be the combined height of the dessert plate and the saucer, in centimeters?

Answer: The dessert plate is 0.93 cm thick, and the saucer is 0.7 cm thick.
Hence their combined height is 0.93 + 0.7 = 1.63 cm.
So the final answer is: 1.63

Question: Kuala Lumpur has 4500 restaurants, Albuquerque has 653 restaurants, Dubai has 13,000 restaurants. How many restaurants do Kuala Lumpur and Dubai have, combined?

Answer: Kuala Lumpur has 4500 restaurants and Dubai has 13,000. So combined they have 4500+13000=17500 restaurants.
So the final answer is: 17500

Question: {question}

Answer:

### Bamboogle compositional CoT prompt.

```
Answer the questions below step by step.
Always conclude with the sentence ’So the
final answer is:’ followed by the final
answer.

Question: What is the capital of the
second largest country in Africa by area?

Answer: The second-largest country
in Africa by area is the Democratic
Republic of the Congo.
The capital of the Democratic Republic of
the Congo is Kinshasa.
So the final answer is: Kinshasa

Question: When did the person who
gave the ’I have a dream’ speech die?

Answer: The person who gave the ’I
have a dream’ speech is Martin Luther King
Jr.
Martin Luther King Jr. died on April 4,
1968.
So the final answer is: April 4, 1968

Question: {question}

Answer:
```

### Bamboogle individual step CoT prompt.

```
Answer the questions below step by step.
Always conclude with the sentence ’So the
final answer is:’ followed by the final
answer.

Question: What is the second largest
country in Africa by area?

Answer: The second largest country
in Africa by area is the Democratic
Republic of the Congo.
So the final answer is: Democratic Republic
of the Congo

Question: When did Martin Luther King Jr.
die?

Answer: Martin Luther King Jr. died on
April 4, 1968.
So the final answer is: April 4, 1968

Question: {question}

Answer:
```

### MultiArith compositional CoT prompt.

```
Answer the questions below step by step.
Always conclude with the sentence ’So the
final answer is:’ followed by the final
numerical answer.

Question: A waiter had 19 customers
to wait on. If 14 customers left and he got
another 36 customers, how many customers
would he have?

Answer: If 14 customers out of 19
left, he would have 19-14=5 customers.
If he got another 36 customers, he would
have 5+36=41 customers to wait on.
So the final answer is: 41

Question: At the fair Kaleb bought 6
tickets. After riding the ferris wheel he
had 3 tickets left. If each ticket cost
9 dollars, how much money did Kaleb spend
riding the ferris wheel?

Answer: Kaleb used 6 - 3 = 3 tickets to
ride the ferris wheel.
Each ticket cost 9 dollars, so Kaleb spent
9 * 3 = 27 dollars riding the ferris wheel.
So the final answer is: 27

Question: {question}

Answer:
```

### MultiArith individual step CoT prompt.

```
Answer the questions below step by step.
Always conclude with the sentence ’So the
final answer is:’ followed by the final
numerical answer.
```

```
Question: A waiter had 19 customers
to wait on. If 14 customers left, how many
customers would he have?

Answer: If 14 customers out of 19
left, he would have 19-14=5 customers.
So the final answer is: 5

Question: Kaleb used 3 tickets to
ride the ferris wheel. If each ticket cost
9 dollars, how much money did Kaleb spend
riding the ferris wheel?

Answer: Each ticket cost 9 dollars
and Kaleb used 3, so Kaleb spent 9 * 3 =
27 dollars riding the ferris wheel.
So the final answer is: 27

Question: {question}

Answer:
```

## E.2 Inference Prompt (Self-ask)

We illustrate below the *self-ask* prompt used to run inference on AgentCoMa. The structure of our prompt is faithful to the original in Press et al. (2023). For consistency with our main setup, we use two examples.

AgentCoMa compositional self-ask prompt.

```
Question: You have a buffet plate that is
1.3 cm thick, a saucer that is 0.7 cm thick,
a dinner plate that is 1.65 cm thick, and
a dessert plate that is 0.93 cm thick. You
want to stack them by decreasing diameter,
the widest at the bottom. What will be the
combined height of the two plates at the
top of the stack, in centimeters?

Are follow up questions needed here:
Yes.
Follow up: Which two plates are at the top
of the stack?
Intermediate answer: For the plates to
be stacked by decreasing diameter, the
stacking order (from bottom to top) is
buffet plate (1.3 cm thick), dinner plate
(1.65 cm thick), dessert plate (0.93 cm
thick), saucer (0.7 cm thick).
Hence the two plates at the top of the
stack are the dessert plate and the saucer.
Follow up: What will be the combined height
of the dessert plate and the saucer, in
centimeters?
Intermediate answer: The combined height
of the dessert plate and the saucer is
0.93 + 0.7 = 1.63 cm.
So the final answer is: 1.63

Question: Kuala Lumpur has 4500
restaurants, Albuquerque has 653
restaurants, Dubai has 13,000 restaurants.
How many restaurants do the Asian cities
have, combined?

Are follow up questions needed here:
Yes.
Follow up: Which cities are Asian between
Kuala Lumpur, Albuquerque and Dubai?
Intermediate answer: Kuala Lumpur and
Dubai are in Asia, whereas Albuquerque is
in America. So the Asian cities are Kuala
Lumpur and Dubai.
Follow up: How many restaurants do Kuala
Lumpur and Dubai have, combined?
Intermediate answer: Kuala Lumpur has
4500 restaurants and Dubai has 13,000.
So combined they have 4500+13000=17500
restaurants.
So the final answer is: 17500

Question: {question}

Are follow up questions needed here:
```

## E.3 Evaluation Prompt

Below, we show the prompt used to elicit LLM-as-a-judge evaluations for non-numerical responses. We use a zero-shot prompt following established literature (Zheng et al., 2023).

LLM-as-a-judge prompt.

```
Please act as an impartial judge and
evaluate the following response to the
question. Determine if the response is
correct given the reference. Respond only
with "yes" if the response is equivalent to
the reference, or "no" if it is not.
Question: {question}
Response: {response}
Reference: {gold_answer}
```

## F API Model Results

Our main results (Appendix A) cover 61 open-weight models. We exclude API models as they cannot be used for interpretability analysis and their undisclosed parameter counts prevent meaningful placement on our size-ordered scales. Nonetheless, we run a preliminary evaluation of OpenAI and Gemini models on AgentCoMa (Table 9). Note that the OpenAI results were obtained on a smaller seed dataset of 60 samples.

| | 1st step | 2nd step | Comp. |
|---|---|---|---|
| GPT-3.5 | 0.68 | 0.60 | 0.28 |
| GPT-4o | 0.72 | 0.76 | 0.68 |
| GPT-4.5 | 0.84 | 0.84 | 0.68 |
| o1 | 0.80 | 0.84 | 0.60 |
| o3-mini | 0.88 | 0.88 | 0.72 |
| Gemini 1.5 Flash 8B | 0.84 | 0.90 | 0.46 |
| Gemini 1.5 Flash | 0.94 | 0.98 | 0.67 |
| Gemini 2.0 Flash | 0.92 | 0.94 | 0.71 |

Table 9: Preliminary results of closed-source models on AgentCoMa.

| **Sample** | **LLM-as-a-judge acc.** |
|---|---|
| *House working* domain | 1.00 |
| *Web shopping* domain | 0.95 |
| *Science experiments* domain | 0.90 |
| *Smart assistant* domain | 0.95 |
| *Travel agent* domain | 0.90 |
| Responses judged *correct* | 0.95 |
| Responses judged *incorrect* | 0.93 |
| Overall | 0.94 |

Table 10: Accuracy of the assessment labels assigned by the GPT-4o judge, with respect to the human ground truths.

## G LLM-as-a-judge Correlation

We use GPT-4o as the judge, as is standard in several recent works (Raju et al., 2024; Zhang et al., 2024; Yehudai et al., 2025b; Zhang et al., 2025b). To measure the correlation between LLM assessment and human judgment, we randomly select 200 non-numerical commonsense step responses generated by different models, stratified across the five AgentCoMa domains as well as across the binary scores assigned by the judge. We have one expert annotator, blind to the LLM judgments, evaluate the responses as 'correct' or 'incorrect' given the ground-truth answers as references. Using these human-assigned labels as the gold standard, we compute the accuracy of the LLM-as-a-judge both overall and within each sub-category. The results, shown in Table 10, indicate that LLM-as-a-judge evaluations are consistently accurate across domains, supporting the reliability of this assessment method.

## H Expert Annotation Guidelines

The samples in AgentCoMa are created by postgraduate-level experts in computing and machine learning. These include the authors and the authors' professional networks. We give them the following instructions, together with examples selected from the first few 'seed' questions created for the benchmark.

*Navigate to the "Data" tab to view some examples of the expected questions.*
*Please create samples only in the following categories:*

- *house working*
- *web shopping*
- *science experiments*
- *smart assistant*
- *travel agent*

*Create questions in each category using each basic operation (add, subtract, multiply, divide).*
*Each created question should require a choice based on commonsense and a single arithmetic operation (i.e. a single addition, or a single subtraction, etc.).*
*The commonsense step must always precede the math step. It needs to be a choice among things.*

| Model Name | Model Type | Bamboogle | | | | MultiArith | | | |
|---|---|---|---|---|---|---|---|---|---|
| | | First step | Second step | Both correct | Composition | First step | Second step | Both correct | Composition |
| Phi4 Mini 3.8B IT | *Instruction-tuned* | 60.8 | 49.6 | 31.2 | 30.4 | 100 | 98.9 | 98.8 | 98.3 |
| Llama3.1 8B IT | | 67.2 | 73.6 | 55.2 | 55.2 | 100 | 97.8 | 97.8 | 99.4 |
| Qwen3 14B | | 72.8 | 64.8 | 52.8 | 47.2 | 100 | 100 | 98.3 | 98.3 |
| Phi3.5 MoE 42B IT | | 77.6 | 75.2 | 66.4 | 66.4 | 100 | 98.9 | 98.9 | 98.9 |
| Llama3.3 70B IT | | 76.0 | 84.8 | 67.2 | 71.2 | 100 | 98.9 | 98.9 | 100 |
| Mixtral MoE 141B | | 77.6 | 81.6 | 67.2 | 68.0 | 100 | 98.3 | 98.3 | 98.3 |
| Art 3B | *Reasoning (SFT)* | 57.6 | 44.8 | 27.2 | 28.0 | 100 | 100 | 97.8 | 97.8 |
| DeepSeek R1 8B | | 63.2 | 45.6 | 42.4 | 36.8 | 98.9 | 96.1 | 95.6 | 93.9 |
| Phi4 Reasoning 14.7B | | 69.6 | 68.8 | 52.0 | 53.6 | 100 | 98.9 | 98.9 | 100 |
| DeepSeekR1 32B | | 72.8 | 68.0 | 53.6 | 53.6 | 100 | 98.9 | 98.9 | 100 |
| Reflection-Llama 70B | | 75.2 | 80.8 | 63.2 | 66.4 | 100 | 98.9 | 98.9 | 92.8 |
| GeneralReasoner 4B | *Reasoning (RL)* | 62.4 | 52.8 | 36.8 | 38.4 | 99.4 | 100 | 98.3 | 98.3 |
| SimpleRL 8B | | 67.2 | 71.2 | 51.2 | 44.8 | 100 | 97.8 | 97.8 | 95.6 |
| GeneralReasoner 14B | | 72.0 | 69.6 | 55.2 | 49.6 | 100 | 97.8 | 97.8 | 98.3 |
| SimpleRL 32B | | 76.0 | 75.2 | 62.4 | 61.6 | 100 | 100 | 98.3 | 98.3 |
| QVQ 72B | | 77.6 | 76.0 | 61.6 | 58.4 | 97.8 | 98.9 | 96.7 | 87.8 |

Table 11: Accuracy scores obtained running 16 LLMs on Bamboogle (Press et al., 2023) and MultiArith (Roy and Roth, 2015). We show the proportions of questions where either step succeeds, both succeed independently, as well as the compositional accuracy. All models are run once using greedy sampling.

*That is, if you were to remove the math part from the question, the question would be a "which" question (see examples).*
*Operations can have more than two terms as long as they do not contain any mixed operators. That is, $3 + 4 + 5$ is fine, $10 - 2 - 3$ is fine, however $3 \times 4 + 2$ is not.*
*Please keep the operations simple enough in terms of number magnitude, the expectation is that both humans and LLMs should be able to do them without using a calculator.*

## I Human Performance Study Details

Human annotators for the performance study are recruited via the Prolific platform[4]. We select non-experts, with the only requirements being fluency in English and high school education. To encourage high-quality answers, we pay annotators a fair compensation of 16.00 USD per hour. Annotators are required to solve all questions in their head or on a piece of paper, without using tools such as calculators or internet search. We report the annotator guidelines below. Note that above each question given to the annotators we provide as examples the same two QA pairs used in the two-shot LLM prompts.

*In this task, you will write down answers to simple questions.*

*The questions can either require simple commonsense reasoning or knowledge (such as knowing that Kuala Lumpur and Dubai are in Asia, but Albuquerque is in North America), or require performing one simple arithmetic operation (either addition, subtraction, multiplication or division), of the kind that can be easily worked out in one's head or on a piece of paper.*

*Some questions could also require BOTH a commonsense step and an arithmetic step. For example, if a question says that "Kuala Lumpur has 4500 restaurants, Albuquerque has 653 restaurants, Dubai has 13,000 restaurants", and asks "How many restaurants do the Asian cities have, combined?", the correct answer involves choosing the Asian cities first (Kuala Lumpur and Dubai) and then adding their restaurants together (4,500 + 13,000 = 17,500).*

*Guidelines:*

*- You must not use Google and you must not use a calculator in this task. Please check your answers – and the arithmetic in them – carefully.*
*- You should not use AI models in this task. AI-generated answers and any answers produced with the aid of unauthorised tools will be discarded and remain unpaid.*
*- We expect high-quality answers. Please ensure your answers are in the correct format.*
*- We provide two examples before each question.*

[4] https://www.prolific.com

*Please structure your answer in a similar manner as the answers in the examples.*
*- Each answer should comprise (very brief) reasoning steps showing how you worked out the solution, and it should end with the sentence "So the final answer is:" followed by the final answer.*
*- For some questions, the final answer will be numerical, for others it will be text-based. Which type of final answer is required will be evident from the question, but also from the example answers above it.*

# J Existing Benchmarks

## J.1 Data Preparation

The Bamboogle and MultiArith benchmarks do not include the underlying reasoning steps as distinct sub-questions. Therefore, we have them extracted manually by expert annotators. The sub-questions are self contained: if information from the first step is required to answer the second step, this information is included within the extracted step.

## J.2 Results

In Table 11 we show the accuracies obtained on Bamboogle and MultiArith. These scores are computed using the same 16 LLMs as in Table 2.

# K Additional Context Analysis

In Table 12, we display LLM accuracies on the first-step (commonsense) sub-questions with extra context. We compute these on the same 16 models shown in Table 2. Average sequence lengths for each (sub-)question group in AgentCoMa are shown in Table 13. In the rest of the section, we show examples of commonsense sub-questions with and without extra context.

**Commonsense sub-question.**

```
You're deep cleaning the kitchen and bedroom, which includes mopping floors where possible. The kitchen floor is tiled, while the bedroom floor has a carpet. Which of the floors can be mopped?
```

**Commonsense sub-question with extra context.**

```
You're deep cleaning the kitchen and bedroom, which includes mopping floors where possible. The kitchen floor is tiled, 3.5 m wide and 4 m long, while the bedroom floor has a carpet and is 4.5 m by 5 m. Which of the floors can be mopped?
```

| Model Name | Model Type | First step (extra context) |
|---|---|---|
| Phi4 Mini 3.8B IT | *Instruction-tuned* | 62.2 |
| Llama3.1 8B IT | | 62.8 |
| Qwen3 14B | | 85.6 |
| Phi3.5 MoE 42B IT | | 91.1 |
| Llama3.3 70B IT | | 91.7 |
| Mixtral MoE 141B | | 93.3 |
| Art 3B | *Reasoning (SFT)* | 62.2 |
| DeepSeekR1 8B | | 82.8 |
| Phi4 Reasoning 14.7B | | 91.1 |
| DeepSeekR1 32B | | 88.9 |
| Reflection-Llama 70B | | 91.7 |
| GeneralReasoner 4B | *Reasoning (RL)* | 73.3 |
| SimpleRL 8B | | 50.6 |
| GeneralReasoner 14B | | 83.9 |
| SimpleRL 32B | | 88.9 |
| QVQ 72B | | 93.3 |

Table 12: LLM accuracies on the first-step (commonsense) sub-questions with extra context. We run the same sample of 16 LLMs shown in Table 2.

| Question type | Avg. seq. length |
|---|---|
| Commonsense | 291.3 |
| Commonsense (extra context) | 427.6 |
| Math | 453.4 |
| Composition | 438.6 |

Table 13: Average sequence lengths for all (sub-)question types in AgentCoMa.

# L Failure Modes

## L.1 Failure Proportions

In Figure 10 we show the difference in failure modes between smaller (8B and below) and larger (14B to 141B) models. We find that for the former category, 58% of the compositional failures occur when the model can solve both individual steps correctly. While this figure is considerable, it is substantially lower than the overall average of 74% shown in Figure 5. This is due to the fact that smaller models generally perform less well on commonsense and math questions. For larger models, however, the proportion of failures where both steps are correct is high (83%). While these capable LLMs excel at the individual steps, their mixed-type compositional reasoning ability remains low.

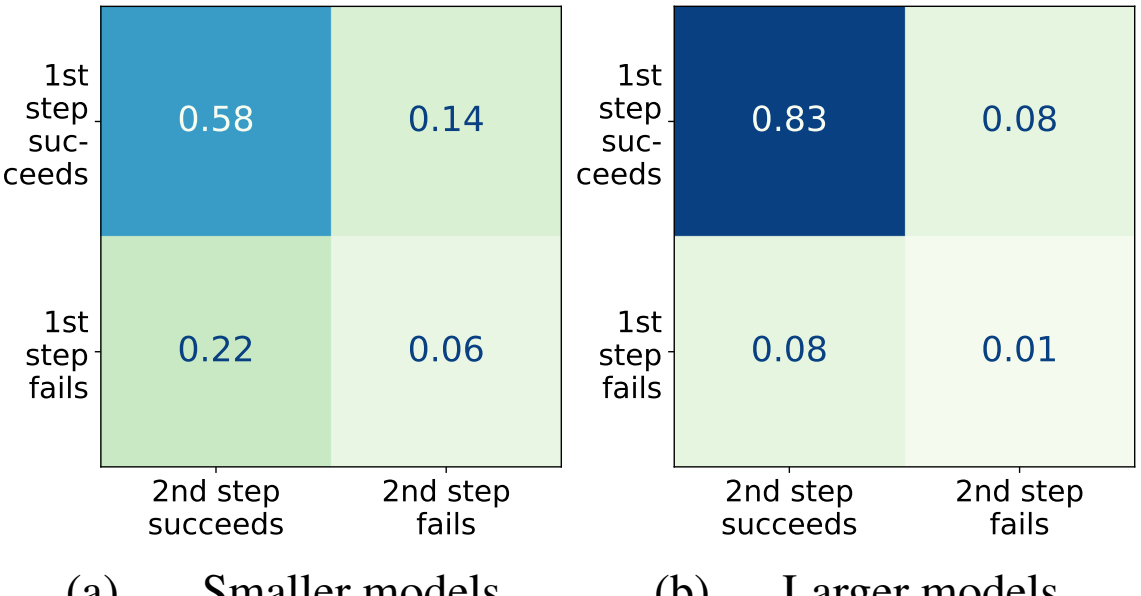

(a) Smaller models (b) Larger models

Figure 10: Proportions of failed compositional questions for smaller models up to 8B (a) and larger ones between 14B and 141B (b). The LLMs are the same ones used for computing the plots in Figure 5.

## L.2 Examples of Failures

We show below an example of compositional failure from the validation set of AgentCoMa, where both individual steps are correct. The responses are generated with Llama 3.3 70B Instruct. We notice that when solving the compositional question, the model suffers from hallucination, i.e., it confuses an individual basketball match with a session containing multiple matches.

**Commonsense-only question:** As a virtual fitness coach, you award a "social" virtual medal for each sport event with at least 20 active athletes that a user participates in. After downloading the app, a user has gone on solo morning jogs, has played semi-professional soccer matches, has played basketball, and has run the London marathon once. Out of these sports events, which had at least 20 active athletes?

**Commonsense-only answer:** 1. Solo morning jogs: These are done alone, so there are no other athletes.
2. Semi-professional soccer matches: These are team sports with 2 teams of 11 players each, so there are at least 22 players in each match.
3. Basketball: This is also a team sport, with 2 teams of 5 players each, so there are at least 10 players in each match.
4. The London marathon: This is a large event with thousands of participants.
Out of these sports events, the semi-professional soccer matches, the London marathon had at least 20 active athletes.
So the final answer is: semi-professional soccer matches and the London marathon

**Math-only question:** As a virtual fitness coach, you award a "social" virtual medal for each sport event with at least 20 active athletes that a user participates in. After downloading the app, a user has gone on a solo morning jog every day for a full week (only one athlete), has played two semi-professional soccer matches (22 athletes on the field), has played basketball four times (12 athletes on the field) and has run the London marathon once (over 50 thousand athletes running). How many social virtual medals has the user accrued so far?

**Math-only answer:** 1. The user has gone on a solo morning jog every day for a full week, but since there is only one athlete, no social medal is awarded.
2. The user has played two semi-professional soccer matches with 22 athletes on the field, so two social medals are awarded.
3. The user has played basketball four times, but since there are only 12 athletes on the field, which is less than 20, no social medal is awarded.
4. The user has run the London marathon once with over 50 thousand athletes running, so one social medal is awarded.
So the total number of social medals is 2 + 1 = 3.
So the final answer is: 3

**Compositional question:** As a virtual fitness coach, you award a "social" virtual medal for each sport event with at least 20 active athletes that a user participates in. After downloading the app, a user has gone on a solo morning jog every day for a full week, has played two semi-professional soccer matches, has played basketball four times and has run the London marathon once. How many social virtual medals has the user accrued so far?

**Compositional answer:** 1. The user has gone on a solo morning jog every day for a full week. Since the user was alone, this does not count as a social event.
2. The user has played two semi-professional soccer matches. Since soccer is a team sport, each match had at least 20 active athletes, so the user gets 2 social medals.
3. The user has played basketball four times. Since basketball is a team sport, each match had at least 20 active athletes (5 players per team, 2 teams per match, and multiple matches per session), so the user gets 4 social medals.

4. The user has run the London marathon once. Since the London marathon has thousands of participants, the user gets 1 social medal.
So the total number of social medals is 2 + 4 + 1 = 7.
So the final answer is: 7

## M Statistical Tests

To assess the statistical significance of differences in lookback attention ratios reported in Section 6.3, we conduct pairwise $t$-tests between all task pairs. As shown in Table 14, all differences are statistically significant at the conventional level ($\alpha = 0.05$). Notably, the math vs composition pair exhibits the largest difference and the highest significance ($p = 0.00001$).

## N Effect of Reasoning Step Order

The compositional tasks in AgentCoMa follow a fixed order: a commonsense step always precedes a mathematical step. To investigate whether this ordering affects the observed compositionality gap, we construct a smaller set of 30 samples where the math step comes first instead. These samples differ from AgentCoMa in an important way: ensuring that the commonsense step involves selecting the correct answer from several possibilities requires performing two or more math operations in the first step, rather than just one. Below, we provide two representative tasks from different domains, along with their CoT-style answers.

| Dataset Pair | *p*-value |
|---|---|
| commonsense vs math | 0.007 |
| commonsense vs composition | 0.007 |
| math vs composition | 0.00001 |

Table 14: Pairwise $t$-test $p$-values for lookback attention ratios across task pairs. All differences are statistically significant ($p < 0.05$), with the math vs composition pair showing the strongest significance.

| First step (math) | Second step (commonsense) | Composition |
|---|---|---|
| 76.7 | 70.0 | 36.7 |

Table 15: Accuracy of Llama 3.1 8B Instruct on reversed-order samples, where the math step precedes the commonsense step. The compositionality gap persists regardless of step order.

**Example Task 1**

Question: Two newly-discovered asteroids are moving toward the Solar System. Both are due to eventually pass exactly by the area where Earth is right now. One, called Atrides, travels at a speed of 1,200,000 miles per day, and is 438,000,000 miles away. The other, called Pharis, travels at a speed of 900,000 miles per day and is 180,000,000 miles away. Which of the two asteroids would likely hit Earth unless it is destroyed?
Answer: The asteroid called Atrides will take 438M/1.2M = 365 days to get to where Earth is right now. The asteroid called Pharis will take 180M/0.9M = 200 days to get to where Earth is right now. Because Earth will be in approximately the same position as it is now in 365 days, but not in 200 days, only Atrides would hit Earth unless it is destroyed. So the final answer is Atrides.

**Example Task 2**

Question: Marika dreams of travelling for the rest of her life. She has two possible plans, and wants to pursue either of them in full. One plan involves travelling to 75 countries and staying 2 years in each. The other involves travelling to all 195 countries and staying 3 months in each. Which plan will she plausibly choose?
Answer: The two plans would take 150 years and 48.75 years, respectively. Since 150 years exceeds the human life span, whatever Marika's current age, the only plausible option is the plan that visits all 195 countries and spends 3 months in each. So the final answer is the second plan.

Table 15 reports the results of Llama 3.1 8B Instruct on this set. The scores are comparable to those on AgentCoMa, suggesting that the compositionality gap persists regardless of step order, within the constraints of this small-scale experiment.